\documentclass[conference]{IEEEtran}
\IEEEoverridecommandlockouts

\usepackage{times}
\usepackage{epsfig}
\usepackage{graphicx}
\usepackage{amsmath}
\usepackage{amssymb}
\usepackage[font=small,labelfont=bf,tableposition=top]{caption}
\usepackage{subfigure}
\usepackage{bbm}
\usepackage{relsize}
\usepackage{cite}
\usepackage{amsfonts}
\usepackage{algorithmic}
\usepackage{textcomp}
\usepackage{xcolor}
\usepackage[T1]{fontenc}
\usepackage{hyperref}

\def\BibTeX{{\rm B\kern-.05em{\sc i\kern-.025em b}\kern-.08em
    T\kern-.1667em\lower.7ex\hbox{E}\kern-.125emX}}

\def\ie{\emph{i.e.~}}
\def\eg{\emph{e.g.~}}

\def\etal{\emph{et al.~}}

\begin{document}


\title{\vspace{1cm}Multi-timescale Trajectory Prediction for Abnormal Human Activity Detection}

\author{Royston Rodrigues \hspace{2cm} Neha Bhargava \hspace{2cm} Rajbabu Velmurugan \hspace{2cm} Subhasis Chaudhuri  \\\\
Department of Electrical Engineering, Indian Institute of Technology Bombay\\
Mumbai, India\\\\

{\tt\small royston.rodrigues@protonmail.com, neha.iitb@gmail.com, rajbabu@ee.iitb.ac.in, sc@ee.iitb.ac.in}
}

\maketitle

\begin{abstract}
   A classical approach to abnormal activity detection is to learn a representation for normal activities from the training data and then use this learned representation to detect abnormal activities while testing. Typically, the methods based on this approach operate at a fixed timescale {---} either a single time-instant (\eg frame-based) or a constant time duration (\eg video-clip based). 
   But human abnormal activities can take place at different timescales. For example, \textit{jumping} is a short term anomaly and \textit{loitering} is a long term anomaly in a surveillance scenario. A single and pre-defined timescale is not enough to capture the wide range of anomalies occurring with different time duration. In this paper, we propose a multi-timescale model to capture the temporal dynamics at different timescales. In particular, the proposed model makes future and past predictions at different timescales for a given input pose trajectory. The model is multi-layered where intermediate layers are responsible to generate predictions corresponding to different timescales. These predictions are combined to detect abnormal activities. In addition, we also introduce an abnormal activity dataset for research use that contains 4,83,566 annotated frames. Our experiments show that the proposed model can capture the anomalies of different time duration and outperforms existing methods.
   \end{abstract}
   Data-set will be made available at \url{https://rodrigues-royston.github.io/Multi-timescale\_Trajectory\_Prediction/}

\section{Introduction}
\label{sec:intro}

Detecting abnormal activities is a challenging problem in computer vision. There is no generic definition available for abnormal events and is usually dependent on the scene under consideration. For example, \textit{cycling} on a \textit{footpath} is typically an abnormal activity whereas it becomes normal on a \textit{road}. To address such a scene context dependency, a typical approach is to consider rare or unseen events in a scene as abnormal. But this may classify the unseen normal activities as abnormal. In general, it may not be possible to know all the normal and abnormal activities during training. We only have access to subsets of normal and abnormal activities. The lack of a generic definition and insufficiency in the data, make it extremely hard for any learning algorithm to understand and capture the nature of abnormal activity. 

More often, the abnormal activity detection problem is posed as an unsupervised learning problem. A common setup of the problem is this - the training data consists of only normal activities and the test data contains normal as well as abnormal activities. A standard approach is to build a model that captures the normality present in the training data. During testing, any deviation from the learned normality indicates the level of abnormality in the test data. Most existing methods formulate it as an outlier detection problem \cite{hasan2016,xu2015,tran2017,TSC-RNN,gan2017}. They attempt to fit the features corresponding to normal activities in a hyper-sphere and the distance of a test feature from this hyper-sphere indicates its abnormality. 

\begin{figure*}
    \centering
    \subfigure{\includegraphics[trim={4cm 0.3cm 0 0}, clip=true,scale=0.29]{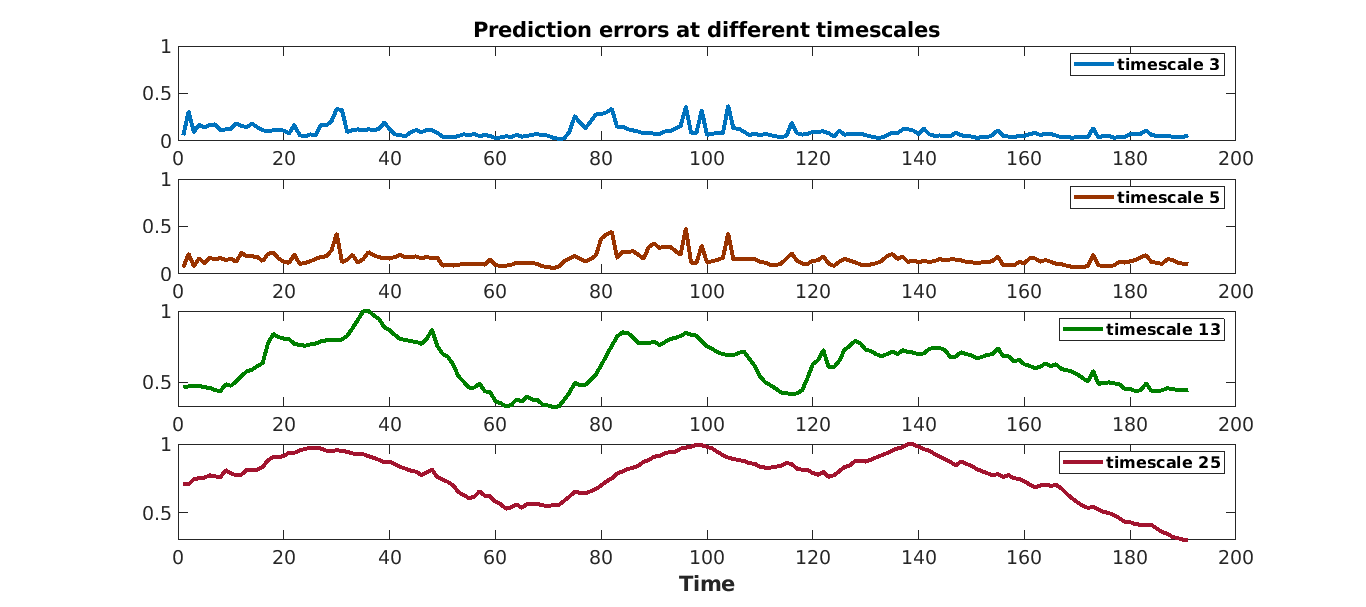}}
    \subfigure{\includegraphics[scale=0.47]{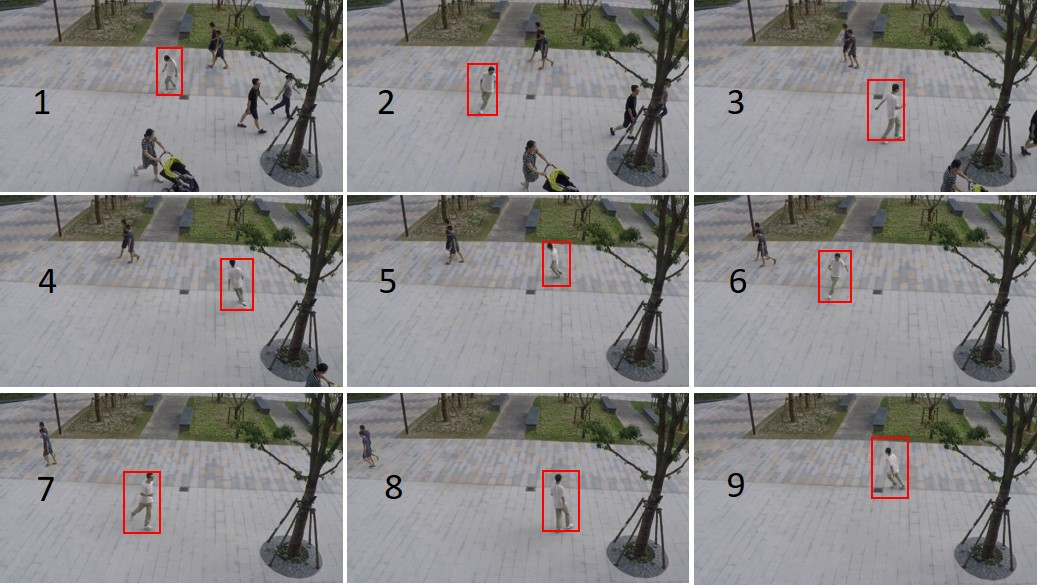}}
    \caption{An illustration of how our model captures a long term anomaly. The anomaly in consideration is \textit{loitering} - the intermediate frames are shown in the right figure with the person involved in red box. The plots in the left shows the prediction errors at different timescales. The prediction errors are less at lower timescales (3 and 5) because at these timescales, the model considers it as a normal activity (\textit{walking}). The errors are higher at larger timescales (13 and 25). At these timescales, the model understands that it is an abnormal pattern of \textit{walking}.}
    \label{fig:loitering_ex}
\end{figure*}

One major limitation of the current methods is that they are trained at a fixed timescale - either a single time-step or a constant number of time-steps. This restricts the model to build understanding of training data at that timescale and hence it may not capture anomalies that occur at other timescales. For example, consider the case of \textit{loitering} - someone wanders at a place for a longer period. The investigation at a smaller time-step may not capture it, because at this timescale it appears as a normal walk. It can be captured only when observed for a sufficient amount of time. Hence, a larger timescale is needed to detect this long-term abnormal activity. Similarly, a larger timescale may not capture the short term anomaly (\eg \textit{jumping}) efficiently. In this paper, we propose a multi-timescale framework to address this problem. The framework has two models - one that makes future predictions and the other one that makes past predictions. They take in a pose sequence to predict future and past sequences at multiple timescales. This is achieved by providing supervision at intermediate layers in the models. For example, the first layer corresponds to a timescale of 3 steps in our setup. At this layer, the input sequence is first broken down into smaller sequences of length three. The model makes 3-step future (and past) predictions for each of these smaller sequences. We combine these predictions together to get a 3-step future (and past) predictions for the input sequence. Similarly, a few other layers in the model generate predictions at different timescales. All the prediction errors from past and future at different timescales are combined appropriately to generate an anomaly score at each time instant in the input sequence. Since the model is trained at different timescales, it tends to learn temporal dynamics at various timescales. An illustration of a long term anomaly (\eg \textit{loitering}) and prediction errors from our model at different timescales is shown in Figure \ref{fig:loitering_ex}. \\

\textbf{Contributions:} We make the following major contributions in the paper:
\begin{itemize}
    \item We propose a bi-directional (past and future) prediction framework that considers an input pose trajectory and predicts pose trajectories at different timescales. Such a framework allows inspection of activities at different timescales (\ie, with different time duration).
    \item We introduce a large dataset that contains a diverse set of abnormal activities. Unlike other datasets, it contains anomalies involving single person to a group of persons. To the best of our knowledge, this is the largest dataset in terms of volume, with 4,83,566 frames.
\end{itemize}

\begin{figure*}
    \centering
    \includegraphics[scale=0.5]{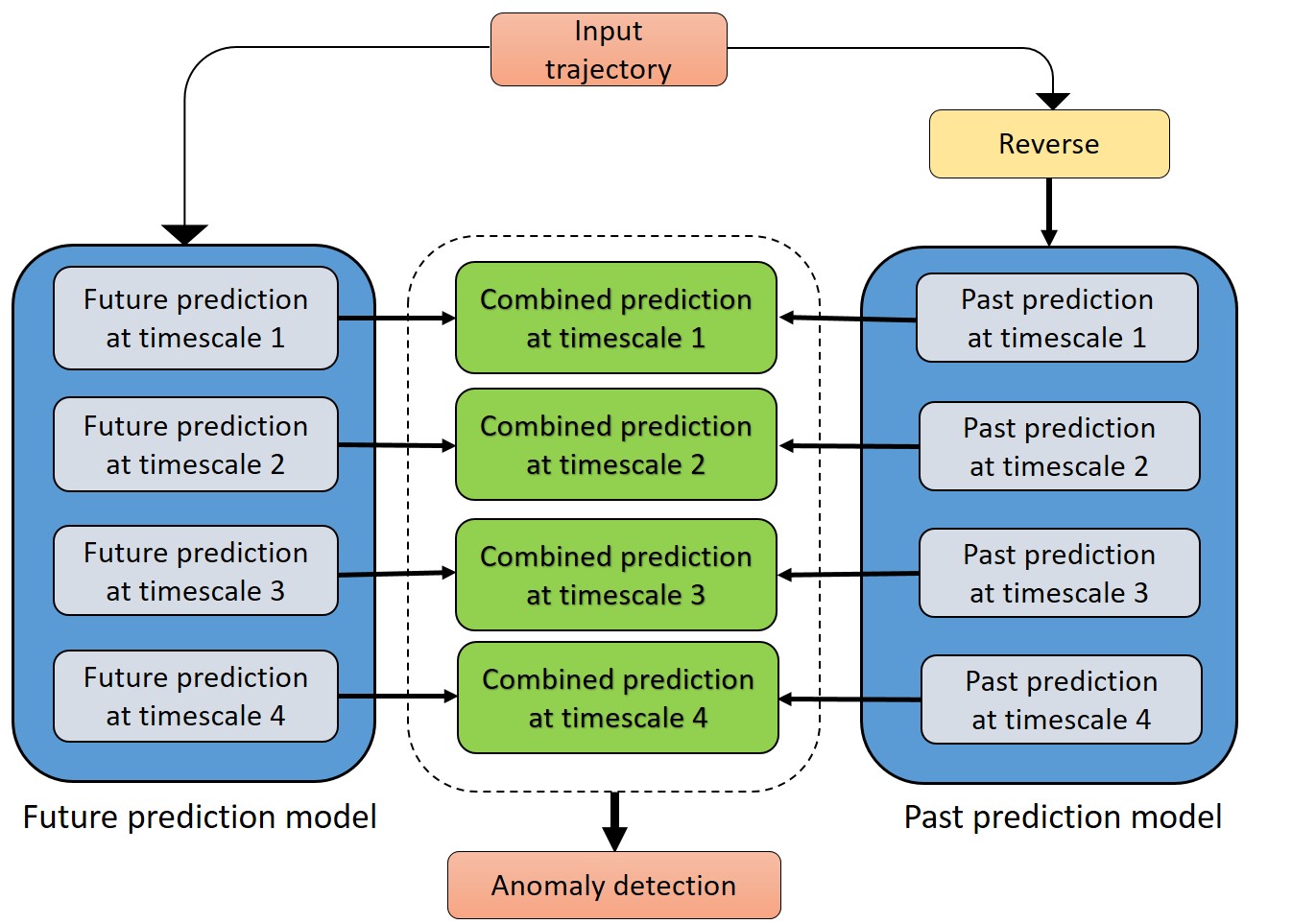}
    \caption{Top-level block diagram of the proposed framework. The future prediction model takes the input sequence and generate predictions at different timescales. To generate predictions at timescale 1, the model first splits the sequence into smaller sub-sequences and then makes future predictions for these sub-sequences. These predictions are combined to get the future prediction for the input sequence at this timescale. To get the past prediction, we reverse the input sequence and pass it to the past prediction model. Finally, all the predictions are combined appropriately to get a final prediction errors for the input sequence that is used to detect abnormal events.}
    \label{fig:block}
\end{figure*}

The paper is organised as follows. The next section reviews the related work in the area of abnormal activity detection. Section \ref{model} details the proposed model. We introduce our dataset in Section \ref{corridor}. We provide the experimental setup, ablation studies, and results in Section \ref{exp}. Finally, the paper concludes in Section \ref{con}.

\section{Related Work}
\label{related}

The problem of human abnormal activity detection has been receiving a lot of interest from computer vision researchers. This is partly because of challenges inherent to the problem and mainly due to its applications. It is interesting to see the evolution of ideas over the years, especially after the introduction of deep learning in this area. In this section, we attempt to summarise this evolution with a few key papers. One of the common and initial approaches was based on reconstruction. Hasan \etal in \cite{hasan2016} used an auto-encoder to learn appearance based representation under reconstruction framework. In \cite{xu2015}, Xu \etal augmented the appearance features with optical flow to integrate motion information. Tran \etal learnt sparse representations by using convolutional winner-take-all autoencoders \cite{tran2017}. Luo \etal in \cite{TSC-RNN} proposed a method based on learning a dictionary for the normal scenes under a sparse coding scheme. To smoothen the predictions over time, they proposed a Temporally-coherent Sparse Coding (TSC) formulation. Ravanbakhsh \etal used GAN to learn the normal representation \cite{gan2017}. In \cite{future}, Liu \etal used GAN with U-net to predict the future frames. Hinami \etal in \cite{hinami2017} proposed a framework that jointly detects and recounts abnormal events. They also learn the basic visual concepts into the abnormal event detection framework. Abati \etal in \cite{novelty} captured surprisal component of anomalous samples, by introducing  an auto-regressive density estimator to learn latent vector distribution of autoencoders. In \cite{unmask}, Ionescu \etal used unmasking to detect anomalies without any training samples.  Recently, Romera \etal in \cite{pose} has proposed a joint framework for trajectory prediction and reconstruction for anomaly detection. One major limitation of the existing methods is that they operate at a single timescale. We take a step forward and propose a multi-timescale model to address the limitations of operating with a single timescale.

\section{Proposed Model}
\label{model}
\begin{figure*}
    \centering
    \subfigure[\textit{Walking}]{\includegraphics[scale=0.6]{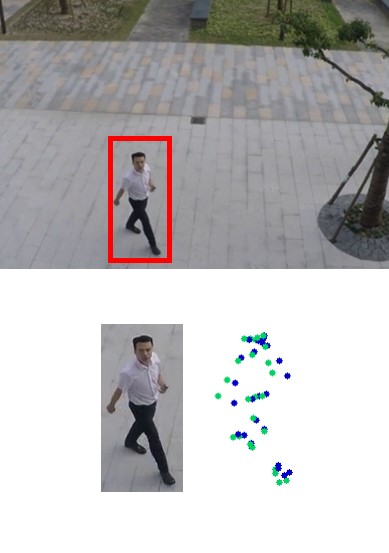}}
    \subfigure[\textit{Cycling}]{\includegraphics[scale=0.6]{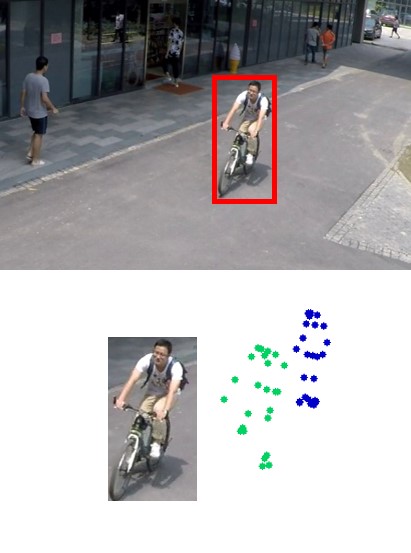}}
    \subfigure[\textit{Leaving bag}]{\includegraphics[scale=0.6]{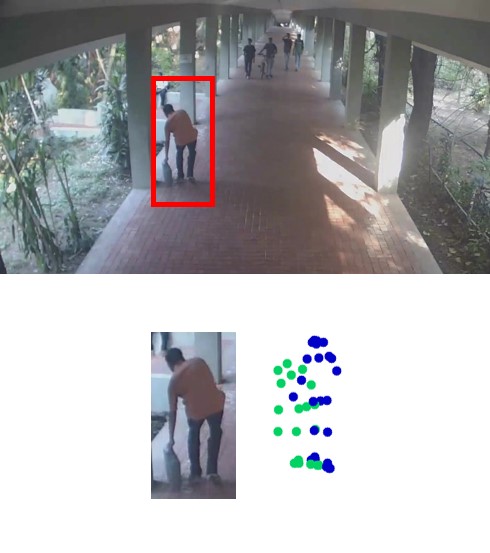}}
    \subfigure[\textit{Throwing bag}]{\includegraphics[scale=0.6]{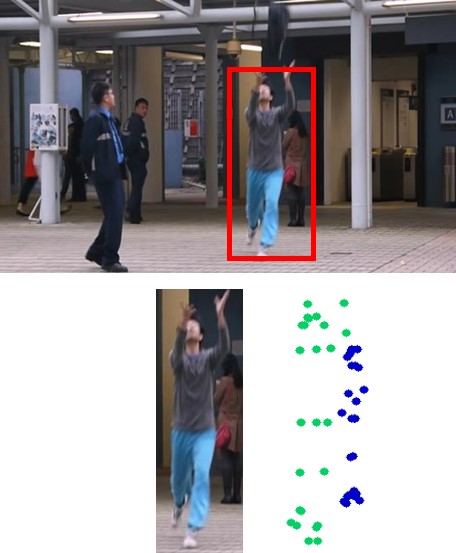}}
    \caption{An illustration of predicted poses from the proposed model. The first scene has a \textit{walking} action while the other three have abnormal actions. The \textit{green} and \textit{blue} poses represent the actual and predicted poses, respectively. The model generate large prediction errors for the abnormal poses.}
    \label{fig:pred_pose}
\end{figure*}

In this section, we present details of the proposed model. To restate, our objective is to develop a framework that is capable of detecting abnormal human activities of different time duration. Keeping this aim in mind, we propose a multi-timescale model that predicts the future trajectories at different timescales. The idea is to develop understanding at different timescales. To further improve the performance, we add an identical model in the framework that interprets the past. We use pose trajectory of human as the input. Besides being compact in nature, the pose trajectory captures the human movements quite well. The top-level illustration of our framework is shown in Figure~\ref{fig:block}. It has two models that make past and future predictions, respectively. At a particular timescale, we combine the predictions from both the models to generate a combined prediction at every time instant. For example, to generate future predictions at timescale 1 (in our setup, timescale 1 represents time duration of 3 steps), the model first splits the sequence into smaller sub-sequences (of length 3) and then makes future predictions (for next 3 steps) for these sub-sequences. These predictions are combined to get the future prediction for the complete input sequence at this timescale. To get the past prediction, we reverse the input sequence and pass it to the past prediction model. Both the models have the same architecture but are trained differently. The past and future predictions at a timescale are combined to get a predicted sequence at that timescale. Finally, all the predictions from different timescales are appropriately combined to get a final prediction error sequence for the input sequence. These prediction error values are compared against a pre-defined threshold. At any time instant $t$, if the error is more than the threshold, the particular time instant is tagged as abnormal. An illustration of our predicted poses is shown in Figure \ref{fig:pred_pose}. It demonstrates higher prediction errors for abnormal activities and low errors for normal activities.

\subsection{Problem setup}
We use human pose trajectory, represented by a collection of 25 points on the human skeleton over a time period, as the input to the model. Let $p^i_j(t)=[x^i_j(t), y^i_j(t)]$ be the image coordinates of $j^{th}$ point in the pose representation of $i^{th}$ person at time $t$. At time $t$, the pose of $i^{th}$ person is represented as $X^i_t = [p^i_1(t), p^i_2(t),\dots, p^i_{25}(t)]^T \in \mathbbm{R}^{50\times1}$ and the corresponding predicted pose is represented by $\hat{X}^{i}_t$. The model takes pose trajectory of a certain length as input and generates predictions at different timescales under a hierarchical framework. For any time instant at any timescale, the model generates multiple predictions because it runs a sliding window over the input signal. These multiple predictions are combined together by averaging to get a final prediction at a particular time instant. Similar approach is adopted to get the past predictions. Finally, the prediction errors from all the timescales are combined to get an abnormality score. In the next sub-section, we discuss the architecture.

\subsection{Model}
\begin{figure*}
    \centering
    \includegraphics[scale=0.6]{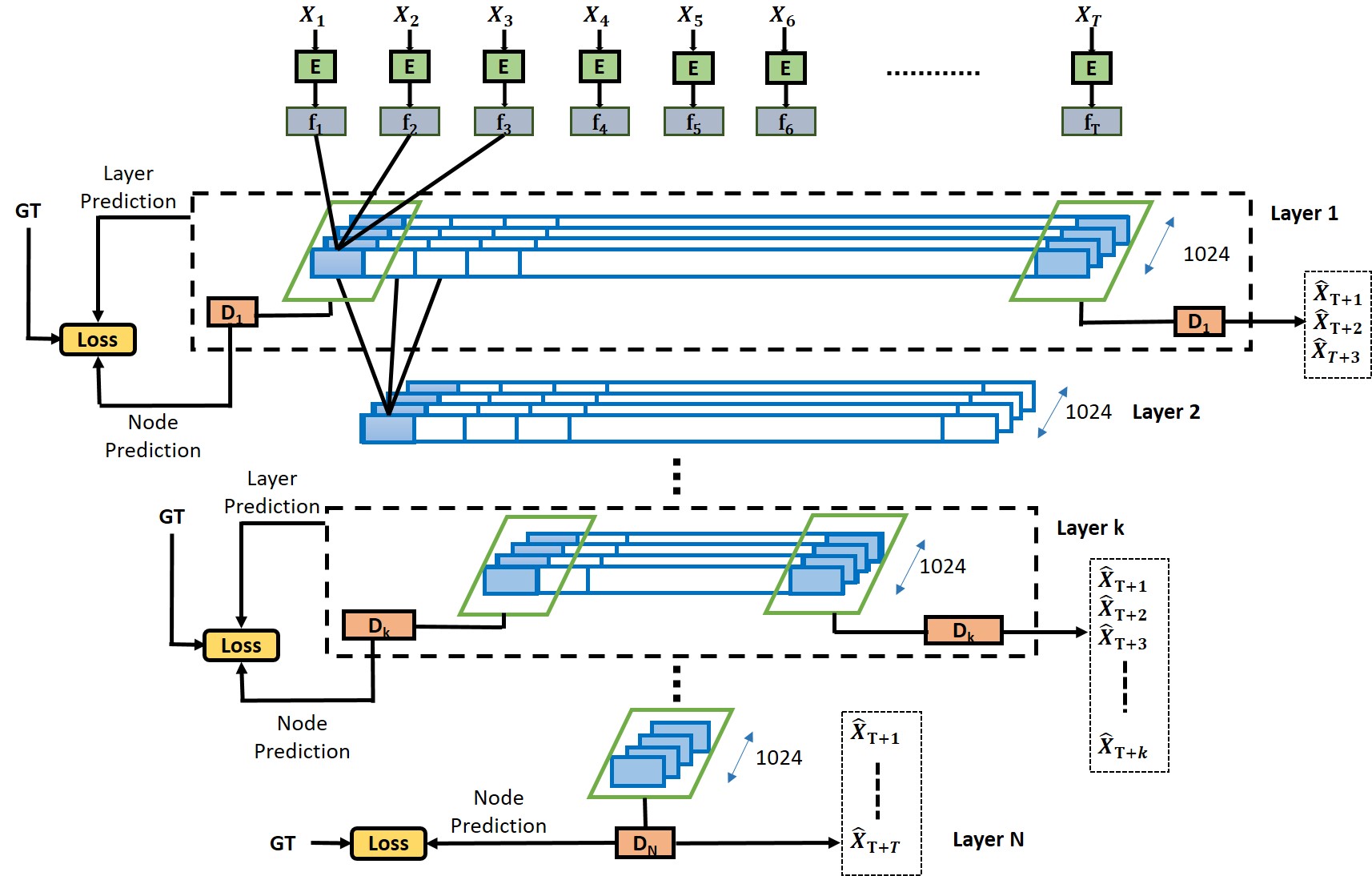}
    \caption{The detailed architecture of the proposed model for future prediction. $X_1$-$X_T$ is the input pose trajectory. After encoding, the vectors $f_1$-$f_T$ are passed to a series of 1D convolutional filters. A few intermediate layers generate predictions at different timescales. To generate predictions, the decoder takes the filtered encoded vector and produces the predictions $\hat{X}$. At these intermediate layers, the node loss and the layer loss are minimised jointly.}
    \label{fig:model}
\end{figure*}

The proposed model is illustrated in Figure \ref{fig:model}. The input to the model is the pose trajectory of an individual $\{X_t\}_{t=1,2,\dots, T}$, where the superscript to identify a person is dropped for better clarity. The pose information at each time instant $X_t$ is passed through an encoder $\mathbf{E}$ to get the corresponding encoded vector $f_t$ of length 1024. These vectors are passed to a series of 1D convolutional filters. Here, a seven layer deep 1D convolutional model is used. We use 1D filters of length 3 for the initial two layers and then we have 1D filters of length 5 for the next five layers. We use 1024 filters at each layer. We train certain intermediate layers to produce predictions at different timescales. This is achieved by providing supervision at these layers. A specific layer corresponds to the $k^{th}$ timescale if its reception field is of length $t_k$ and is responsible to generate $t_k$ future (and past) steps in the trajectory. To provide supervision, we train a decoder to predict the future sequence. In our setup, we provide supervision to the layers corresponding to the timescales - 3, 5, 13 and 25. The decoder $\mathbf{D_k}$ at time scale $t_k$ consists of two fully-connected layers, of length 1024 and $t_k*50$. The encoder $\mathbf{E}$ consists of two fully-connected layers of length 1024 to transform frame level pose features of length 50 to 1024. The encoder $\mathbf{E}$ operates at an individual frame level. Encoded 1024 dimension pose features $f_t$  are stacked to form a time series corresponding to the pose trajectory under consideration. The output length for the decoder depends on the specific timescale. In the next sub-section, we discuss the loss function in detail.

\subsection{Loss function}
To compute the total loss of the model during training, we add the losses generated by the intermediate layers where we provide supervision. At such a layer, we have two types of losses - one at a node level, and another at the layer level. The loss at a node level computes the prediction error between the predictions generated by a node and the corresponding ground truth. The loss at a layer level computes the total prediction error generated by the layer for a complete input sequence. Since, there are multiple predictions generated at a particular time instant by different nodes, we use a sliding window approach to calculate the total loss at a layer. In particular, we take the average of all the prediction errors generated at a time instant by different nodes to compute the total prediction error at a particular time instant. The average prediction errors at all the time instants are added to get the layer loss. The total loss at the $j^{th}$ layer with $M_j$ number of nodes is given as, 
\begin{equation}
    \mathbbm{L}_j = \mathlarger{\sum}_{i=1}^{M_j}L^j_1(i) + \mathlarger{\sum}_{t=1}^{T}L^j_2(t),
    \label{eq:total_layer_loss}
\end{equation}
where $L^j_1(i)$ is the loss for $i^{th}$ node and $L^j_2(t)$ is the loss at time $t$ in the $j^{th}$ layer. The first term corresponds to the total node loss and the second term corresponds to the total layer loss at $j^{th}$ layer. Let the $i^{th}$ node in $j^{th}$ layer make predictions for the duration $\mathbbm{T}(i)=[t_{si},~t_{ei}]$ and generate prediction error $\textit{e}(t,i)$ for a particular time instant $t \in [t_{si},~t_{ei}]$. The $i^{th}$ node loss is computed as follows:
\begin{equation}
\label{eqn:loss_l1}
    L^j_1(i) = \mathlarger{\sum}\limits_{t=t_{si}}^{t_{ei}}\textit{e}(t,i)
\end{equation}

\begin{figure*}
    \centering
    \includegraphics[scale=0.65]{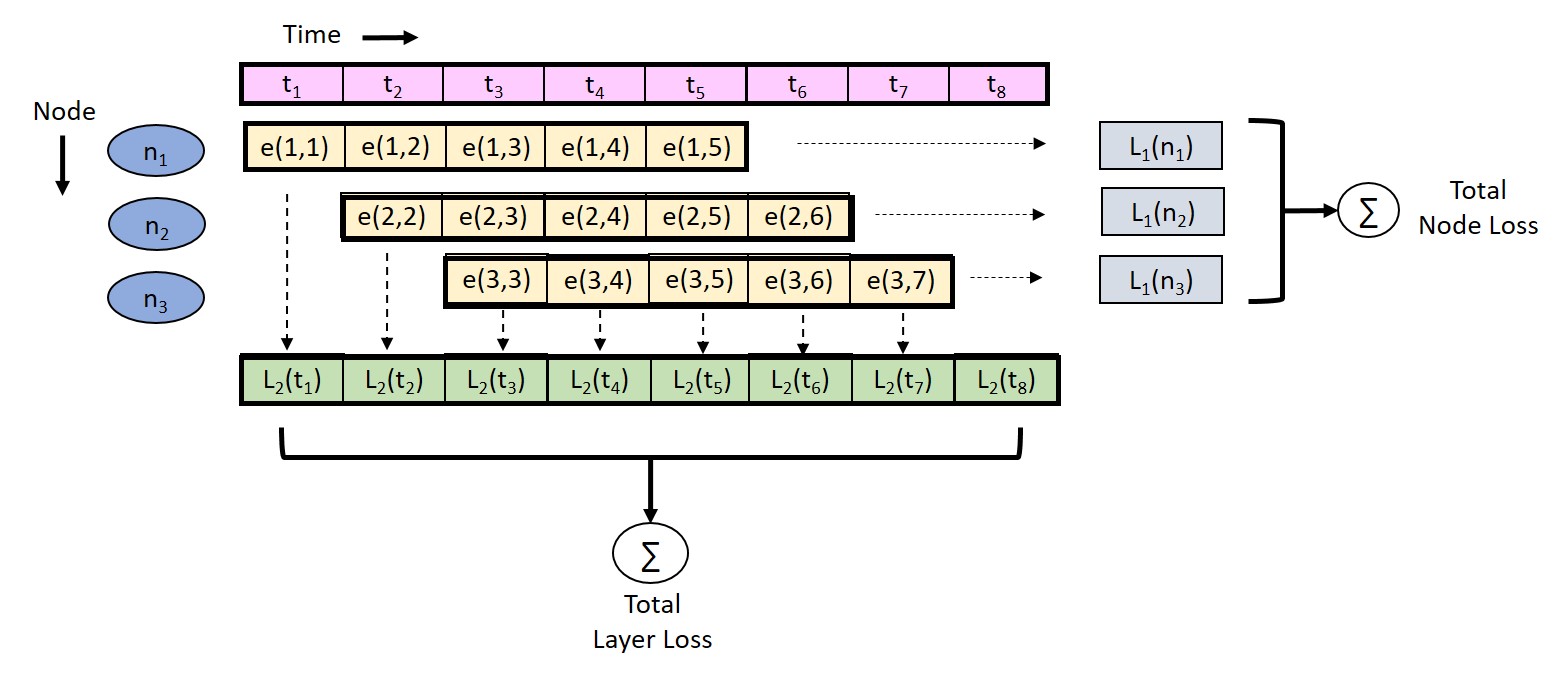}
    \caption{Computation of node and layer losses at a particular timescale. To compute the prediction loss generated by the node $n_1$, we simply add the errors $e(1,1)$ to $e(1,5)$ because $n_1$ makes predictions for $t_1$ to $t_5$. We add all the node losses to get the total node loss. To compute the layer loss, we first find the average loss at each time instant that is contributed by multiple nodes \ie, $L_2(t_1)$ to $L_2(t_8)$. For example, to compute $L_2(t_2)$, we take weighted average of $e(1,2)$ and $e(2,2)$. We add these average errors to get the total layer loss.}
    \label{fig:loss}
\end{figure*}

To compute the layer loss, we simply take the average of prediction errors generated by different nodes for a particular time instant. We finally add these average errors for all the time instants to get the layer loss.
\begin{equation}
\label{eqn:loss_l2}
    {L}^j_2 = \frac{\sum\limits_{i=1}^{M_j}\textit{e}(t,i)\mathbbm{1}(t \in \mathbbm{T}(i))}{\sum\limits_{i=1}^{M_j}\mathbbm{1}(t \in \mathbbm{T}(i))}
\end{equation}
A simple example to illustrate the loss computation at a layer is shown in Figure \ref{fig:loss}. The total model loss for a model with $N_{ts}$ number of timescales is,
\begin{equation}
    {Loss} = \mathlarger{\sum}_{j=1}^{N_{ts}}\mathbbm{L}_j
\end{equation}
We use weighted mean square error (\textit{mse}) to compute the prediction error. 
\begin{equation}
    \textit{e} = \sum_{k=1}^{25}w_k{(\hat{p}_k - p_k)^2},
    \label{eq:weight_mse}
\end{equation}
where $p_k$, $\hat{p}_k$ are the original and predicted $k^{th}$ pose points, respectively. The weight $w_k$ is obtained from the confidence of pose estimator \cite{openpose} for the $k^{th}$ point as
\begin{equation}
    w_k = \frac{c_k}{\mathlarger{\sum}_{i=1}^{25}c_i},
\end{equation}
where $c_k$ is the confidence score for the $k^{th}$ point given by the pose detector.

With the loss function in Eq. (\ref{eq:total_layer_loss}), we jointly minimise the local loss at a node and the global loss at a layer. Minimising the node loss makes the prediction better at a node level whereas minimising the layer loss drives the nodes to interact with each other to reduce the layer loss. Another advantage of using the layer loss under a sliding window approach during training is that it simulates the testing scenario. During testing, it is common to use a sliding window over a long test sequence to get an input sample of suitable length for the model.

\subsection{Anomaly detection}
\label{pred_error}
In this section, we discuss the method for anomaly detection during testing. The trained model predicts pose trajectories for a human at different timescales. The past and future prediction errors produced at these different timescales are combined using a voting mechanism to compute the final prediction error. At any time instant $t$, the errors are combined as follows: 
\begin{equation}
    error(t) = \frac{\sum_{j\in S}L^j_2(t)}{|S|},
\end{equation}
where S is a set of timescales that contains predictions for the time instant $t$ and $L^j_2(t)$ is the $j^{th}$ layer loss at time $t$, as in~(\ref{eqn:loss_l2}). Note that at any time instant, there can be more than one human resulting in multiple error plots - one for each human. In such a case, we take the maximum of prediction errors among all the individuals, at a time instant. That is, 
\begin{equation}
    error(t) = \frac{\sum_{j\in S}\max\{L^j_2(t,p_k), \forall k\}}{|S|},
\end{equation}
where $L^j_2(t,p_k)$ is the $j^{th}$ layer loss at time $t$ for the $k^{th}$ individual. During testing, $error(t)$ is compared against a threshold. If it exceeds the threshold, then the time instant $t$ is tagged as abnormal.

\begin{table*}
\centering
\def\arraystretch{1.25}
\begin{tabular}{ |c | c c c c c| } 
 \hline
  & HR-ShanghaiTech & ShanghaiTech & HR-Avenue & Avenue & \textit{Corridor}\\ 
  \hline
 Conv-AE \cite{hasan2016} & 69.80 & 70.40 & 84.80 & 70.20 & -\\
 Ionescu \etal \cite{unmask} & - & - & - & 80.60 & - \\
 TSC-rRNN \cite{TSC-RNN} & - & 68.00 & - & 81.71 & -\\ 
 Liu \etal \cite{future} & 72.70 & 72.80 & 86.20 & $\mathbf{84.90}$ & 64.65\\ 
 Abati \etal \cite{novelty} & - & 72.50 & - & - & -\\
 Morais \etal \cite{pose} & 75.40 & 73.40 & 86.30 & - & 64.27\\ 
 \hline
 Ours & $\mathbf{77.04}$ & $\mathbf{76.03}$ & $\mathbf{88.33}$ & 82.85 & $\mathbf{67.12}$\\
 \hline
\end{tabular}
\vspace{10pt}
\caption{Performance comparison with the existing techniques.}
\label{table:performance}
\end{table*}
\section{Corridor dataset}
\label{corridor}
We provide a brief introduction of the proposed \textit{Corridor} dataset for abnormal human activity. The videos are captured in a college campus. The scene consists of a corridor where the normal activities are usually \textit{walking} and \textit{standing}. We enacted various abnormal activities with the help of volunteers. The dataset contains variety of activities and has single person to group level anomalies. The annotations for normal and abnormal are provided at frame level. To the best of our knowledge, this is the largest dataset in terms of volume. A comparison of the proposed dataset with other datasets is given in Table \ref{table:iitb}. The last column mentions the abnormal activities present in the datasets. Out of 1,81,567 test frames, the number of abnormal frames is 1,08,278. Additionally, we provide the class of each abnormal activity that can be used for classification purposes. Table \ref{table:iitb} shows sample images from the dataset corresponding to all the activities. To keep the privacy of volunteers intact, we have blurred the faces in the images. We used the algorithm proposed by \cite{face} to detect the faces. In the next section, we discuss our experimentation in detail.

\section{Results and Discussions}
\label{exp}
In this section, we discuss our experimental setup and results. We tested our method on the proposed dataset and two public datasets namely, ShanghaiTech Campus dataset \cite{TSC-RNN} and Avenue \cite{Avenue}. The ShanghaiTech Campus dataset contains videos from 13 different cameras around the ShanghaiTech University campus. A few examples of human anomalies present in the dataset are \textit{running}, \textit{fighting}, \textit{loitering}, and \textit{jumping}. Avenue dataset is captured at CUHK campus. It contains anomalies such as \textit{throwing a bag}, \textit{running}, \textit{walking near the camera}, and \textit{dancing}. The training set has 16 videos and the test set has 21 videos. Since we are interested in human anomaly, similar to \cite{pose}, we test our algorithm primarily on HR-ShanghaiTech and HR-Avenue datasets, proposed by them. In this section, we first discuss our pre-processing to generate the pose trajectories, followed by training and testing schemes. We then compare the performance of our model with the state-of-the-art methods.

\subsection{Data preparation}
In this section, we discuss our method to obtain the pose trajectories from the videos. We first run a human detector \cite{detector} on all the videos. We obtain bounding box trajectories from the detections using the multi-target tracker proposed by \cite{mtt}. Finally, we run a pose-detector \cite{openpose}, \cite{openpose_RT} on these trajectories to get the pose trajectories. This pose detector also produces confidence values for each of 25 points. We use these values in Eq. (\ref{eq:weight_mse}) to calculate the weighted \textit{mse}.

\subsection{Training scheme}
In this section, we discuss our training paradigm. To generate predictions at different timescales, we provide supervision at pre-chosen layers in our multi-layered model. In our model, we provide supervision to the layers - 1, 2, 4, and 7 corresponding to timescales of 3, 5, 13, and 25, respectively. Each

\begin{figure*}
    \centering
    \subfigure[Chasing]{\includegraphics[scale=0.06]{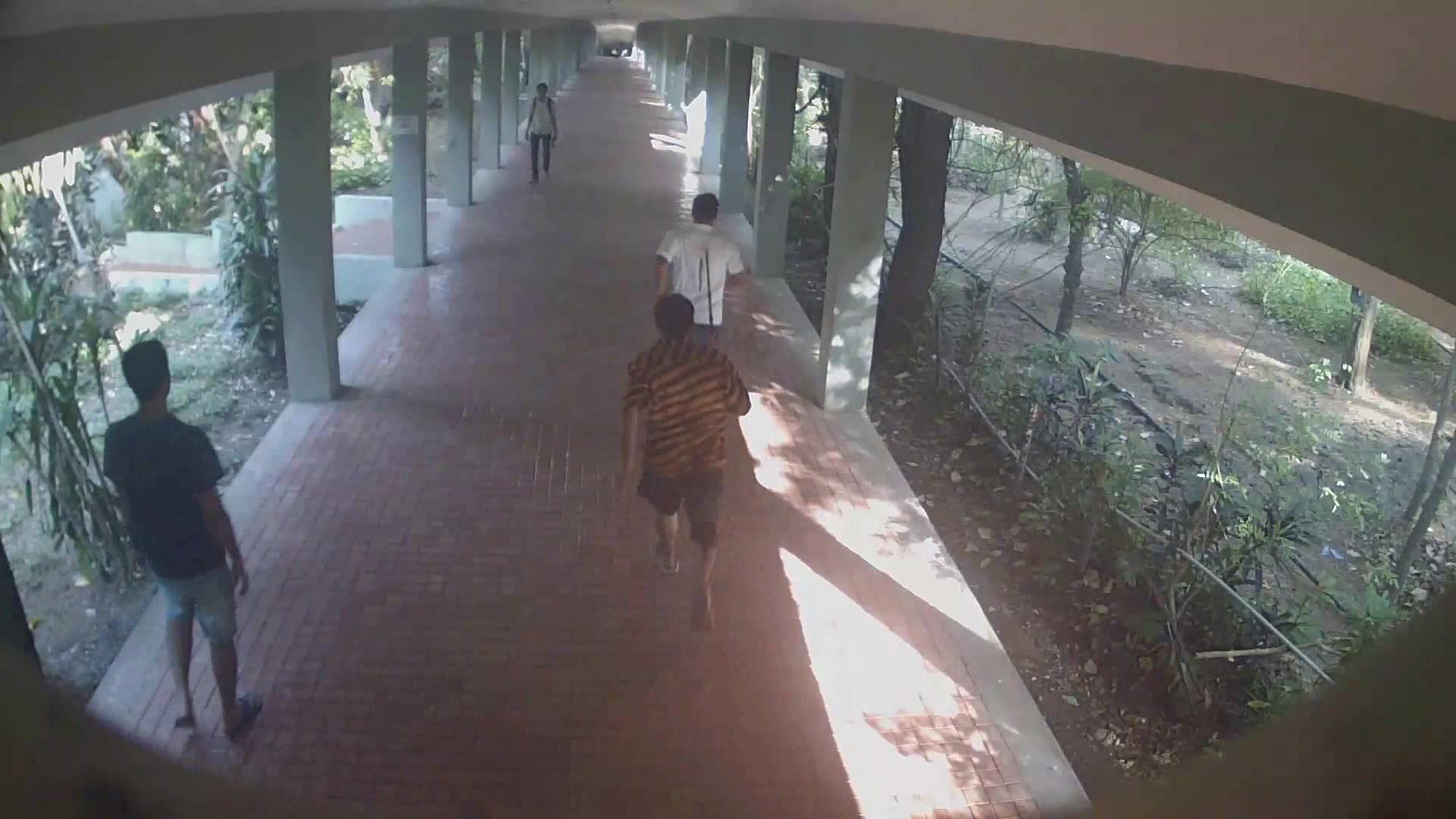}}
    \subfigure[Suspicious object]{\includegraphics[scale=0.06]{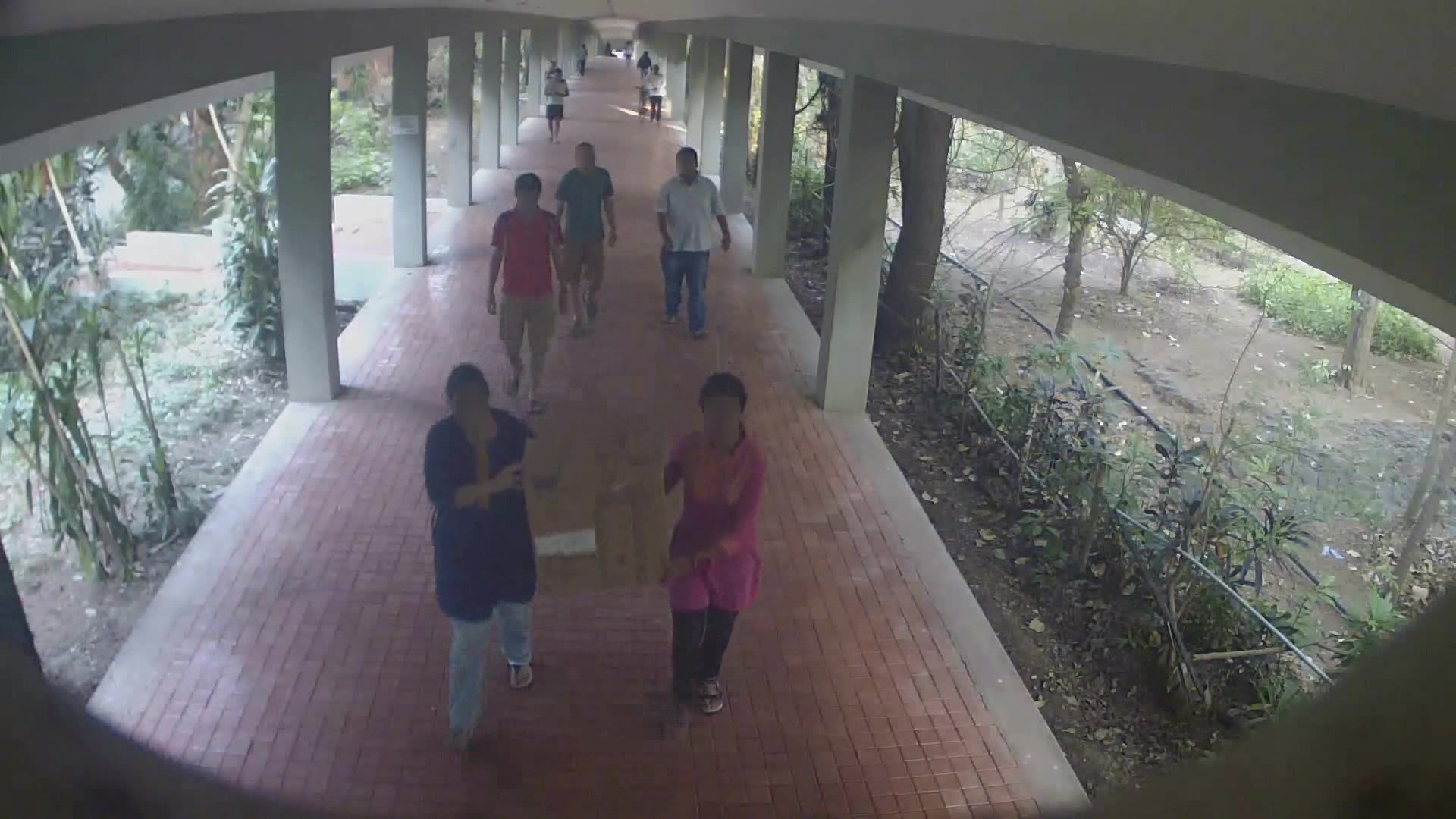}}
    \subfigure[Cycling]{\includegraphics[scale=0.06]{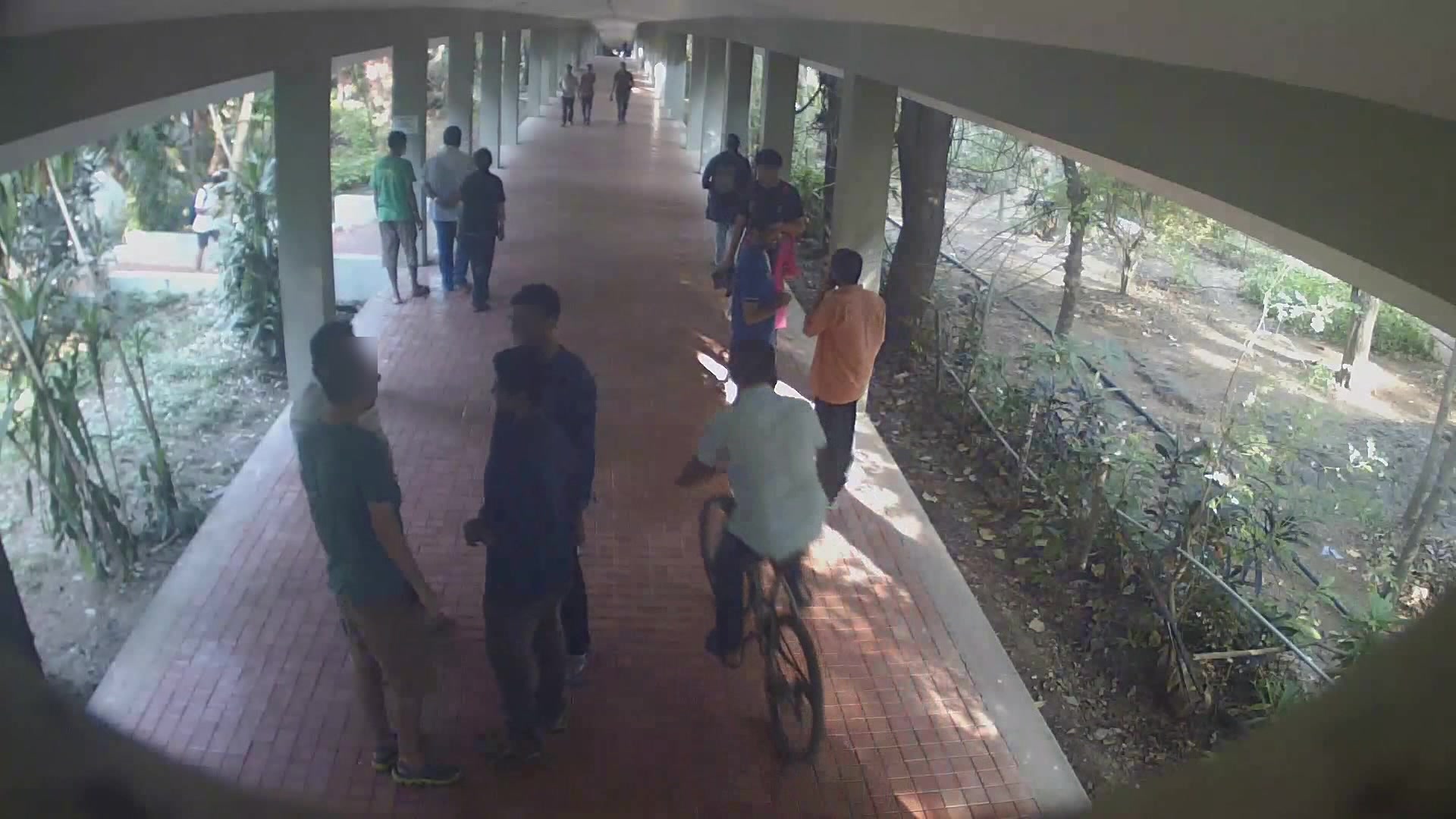}}
    \subfigure[Hiding face from camera]{\includegraphics[scale=0.06]{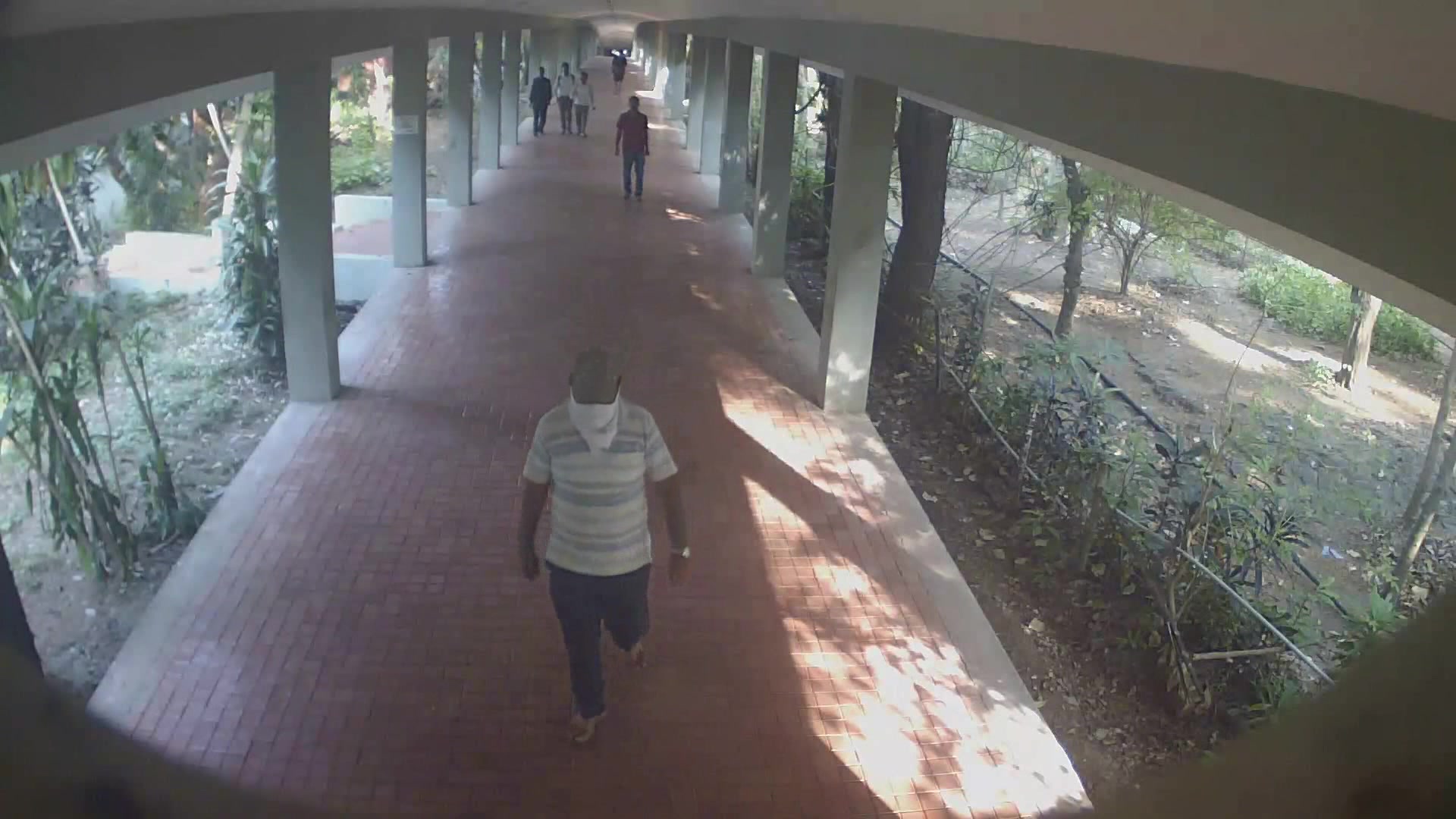}} \vspace{-0.2cm}\\
    \subfigure[Fighting]{\includegraphics[scale=0.06]{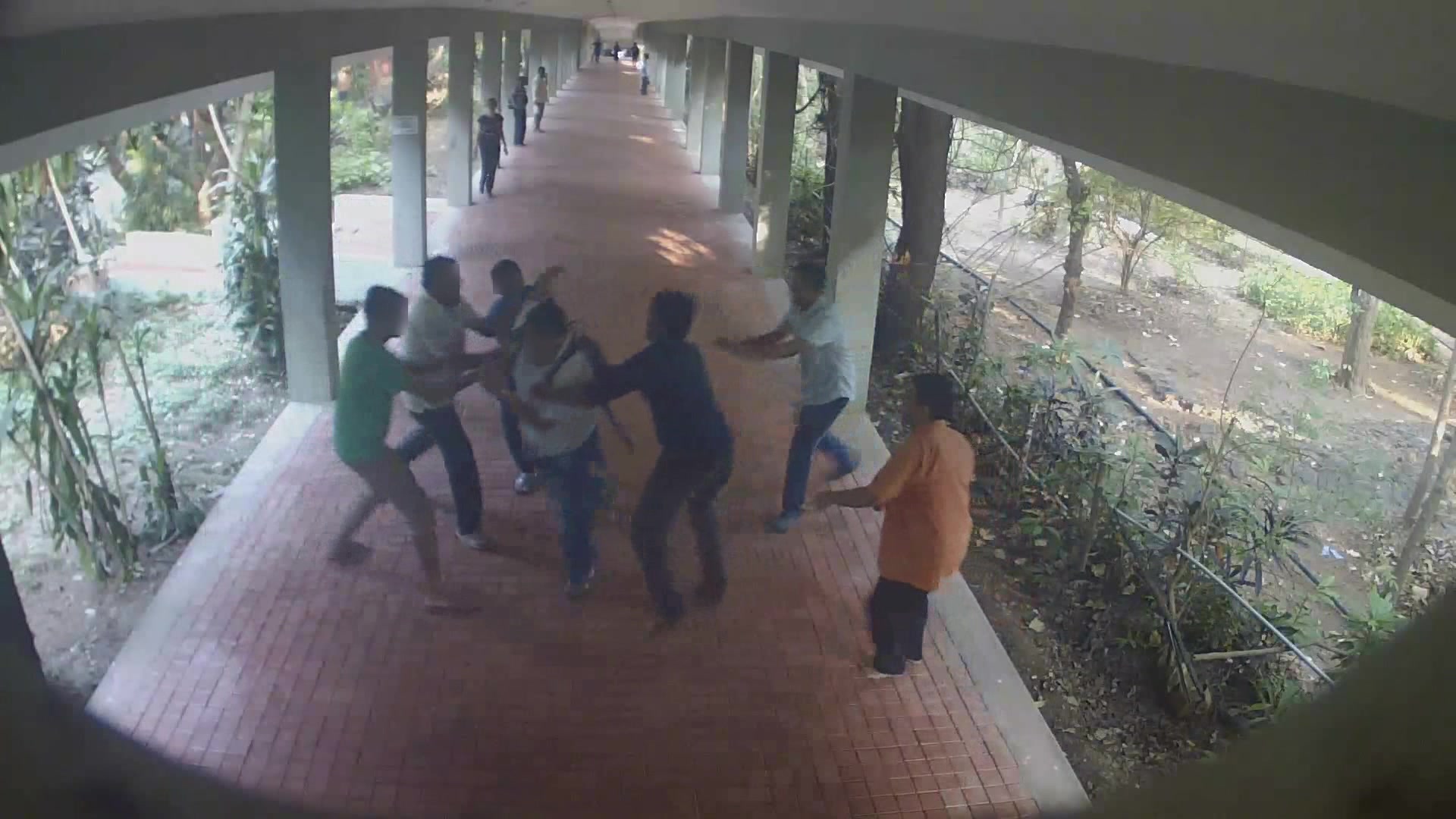}}
    \subfigure[Loitering]{\includegraphics[scale=0.06]{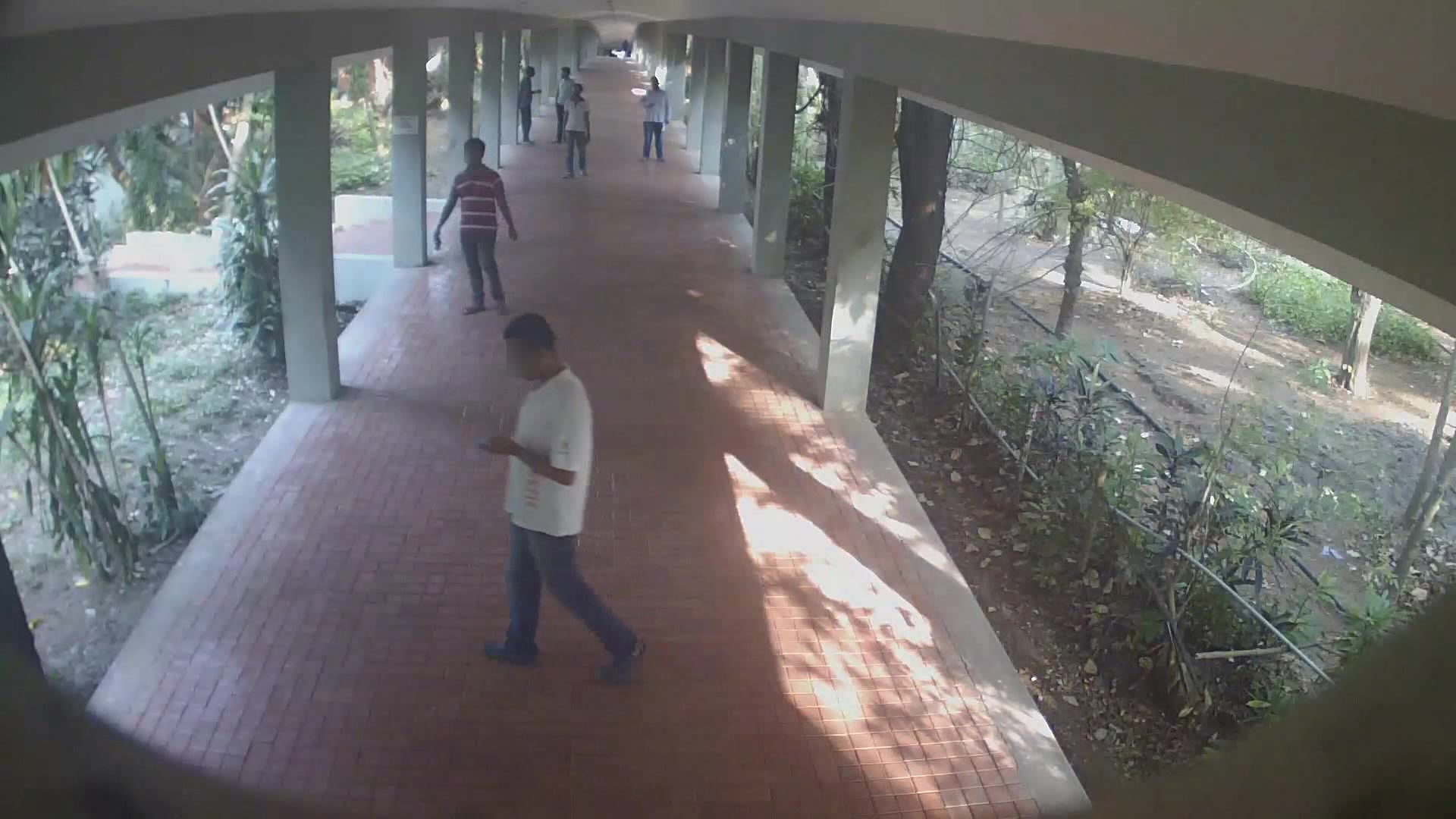}}
    \subfigure[Playing with ball]{\includegraphics[scale=0.06]{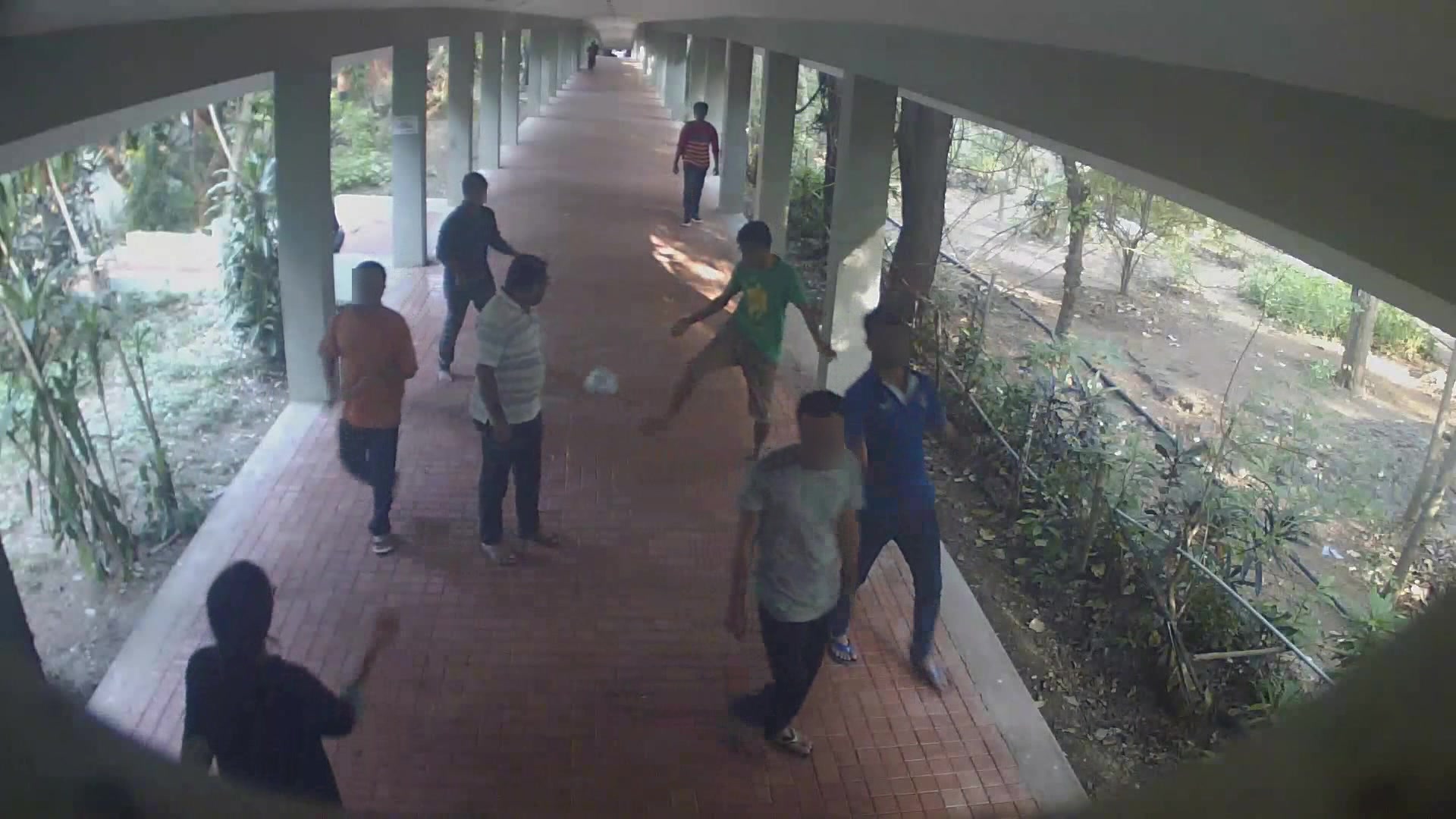}}
    \subfigure[Protest]{\includegraphics[scale=0.06]{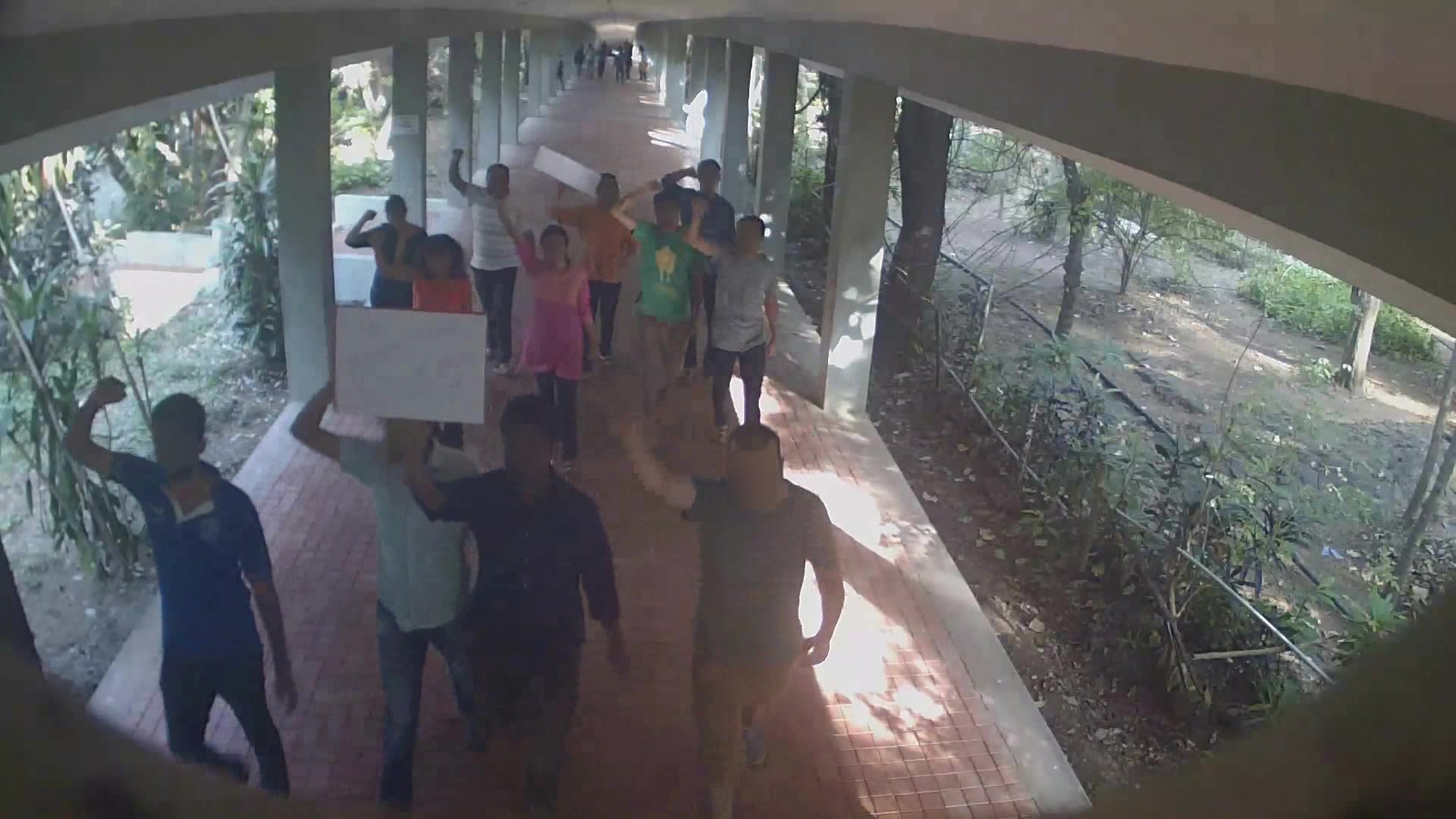}}\vspace{-0.2cm}\\
    \subfigure[Sudden running]{\includegraphics[scale=0.06]{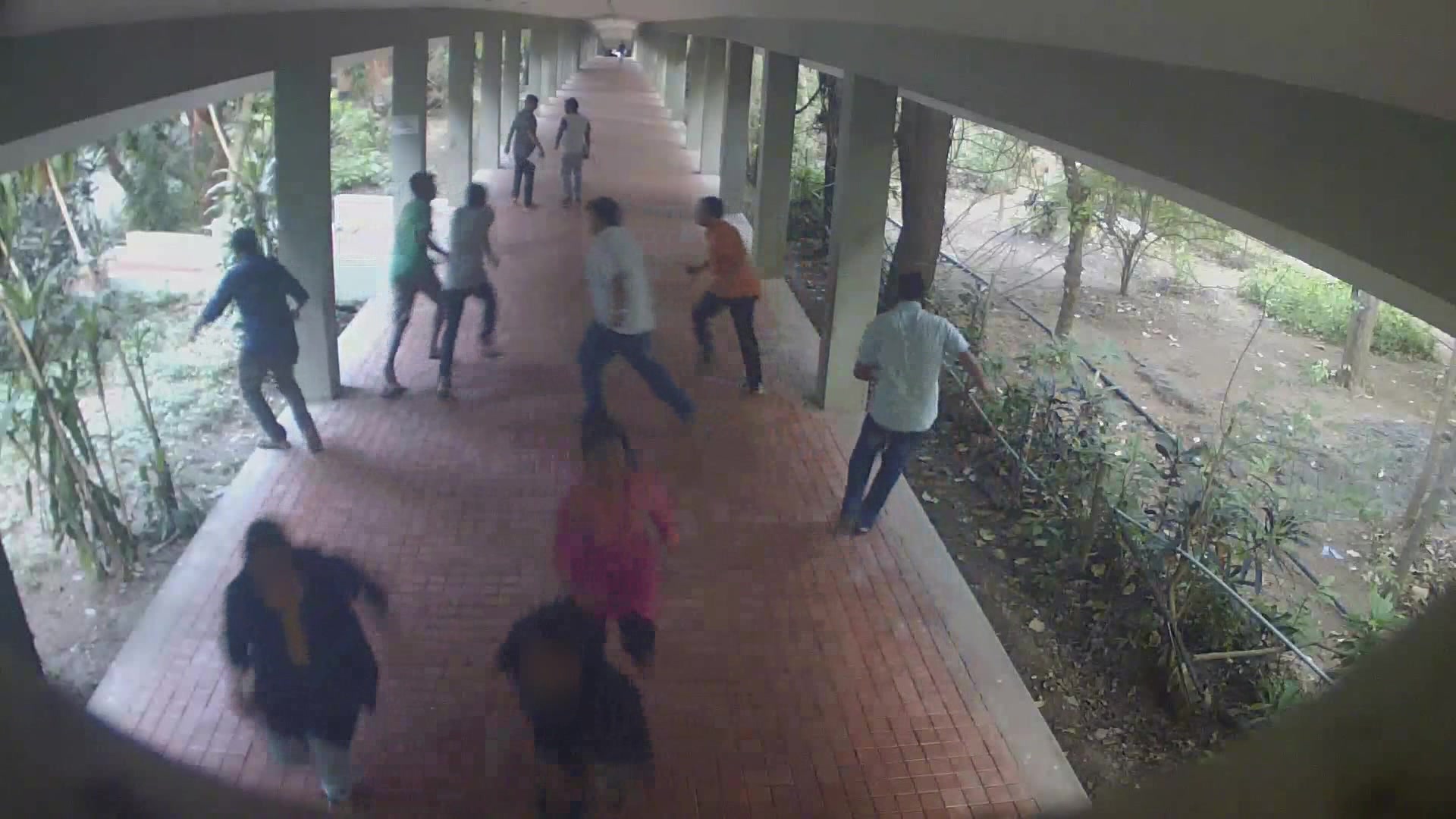}}
    \subfigure[Unattended baggage]{\includegraphics[scale=0.06]{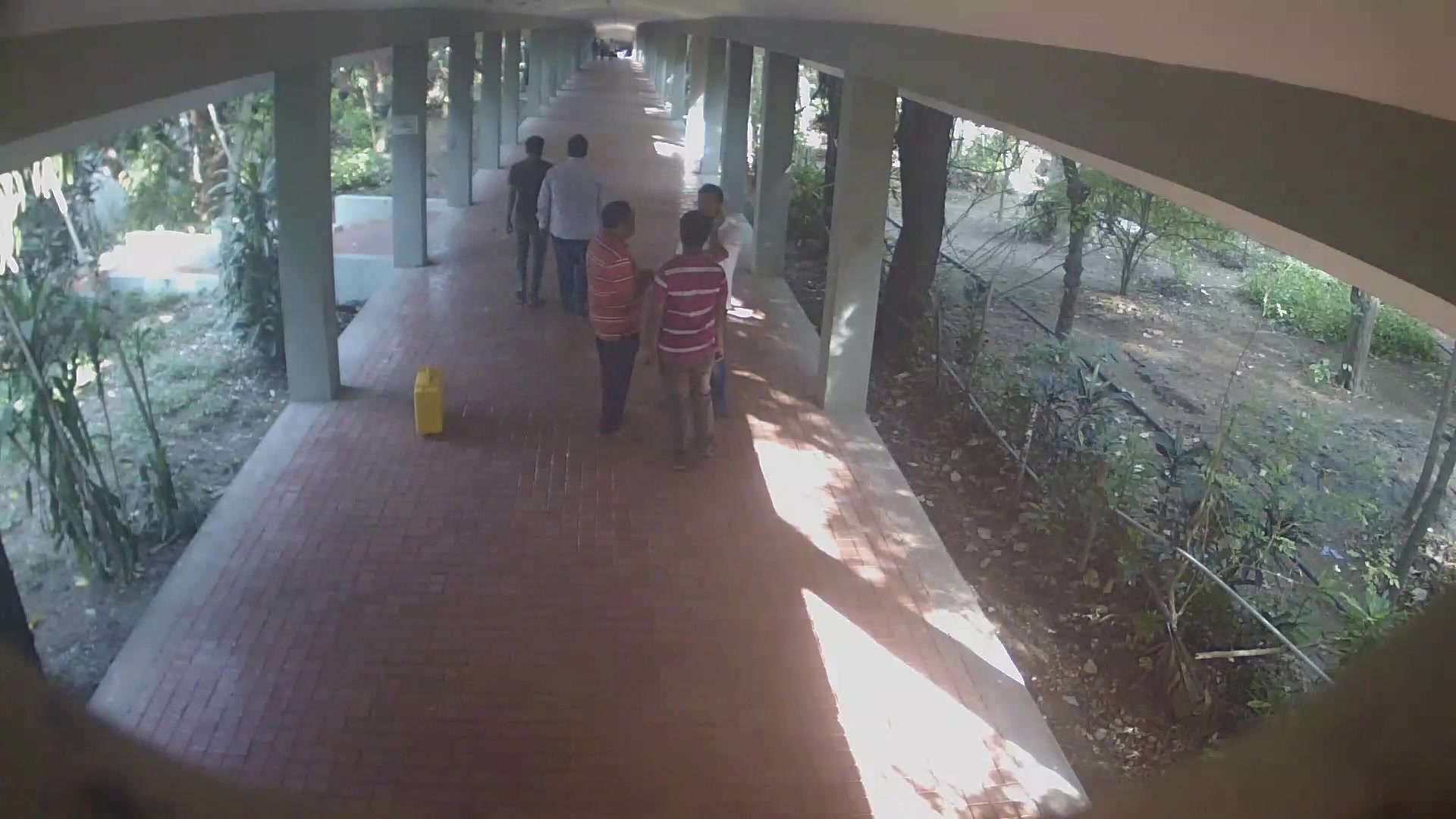}}
    \subfigure[Normal scene]{\includegraphics[scale=0.06]{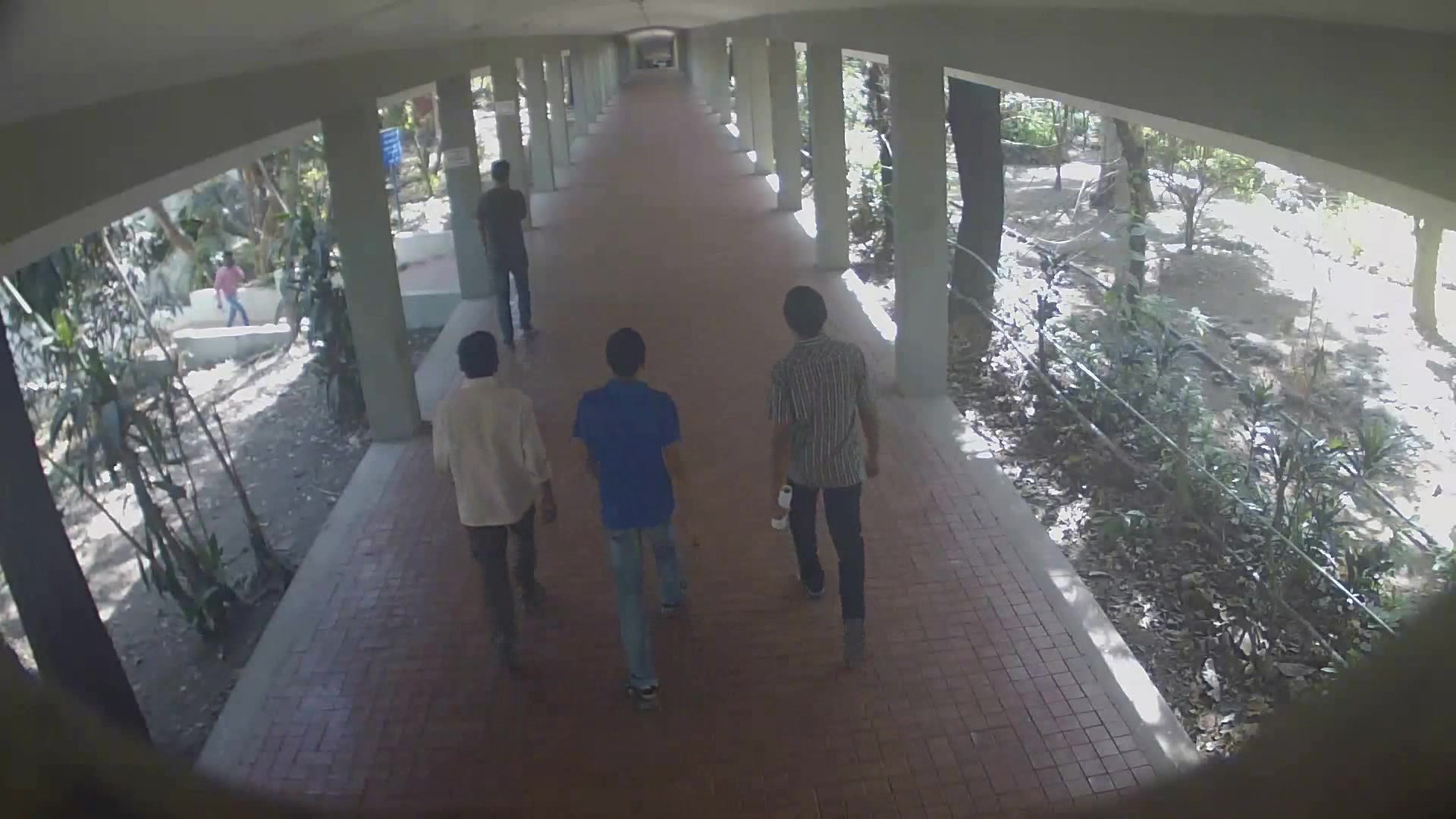}}
    \subfigure[Normal scene]{\includegraphics[scale=0.06]{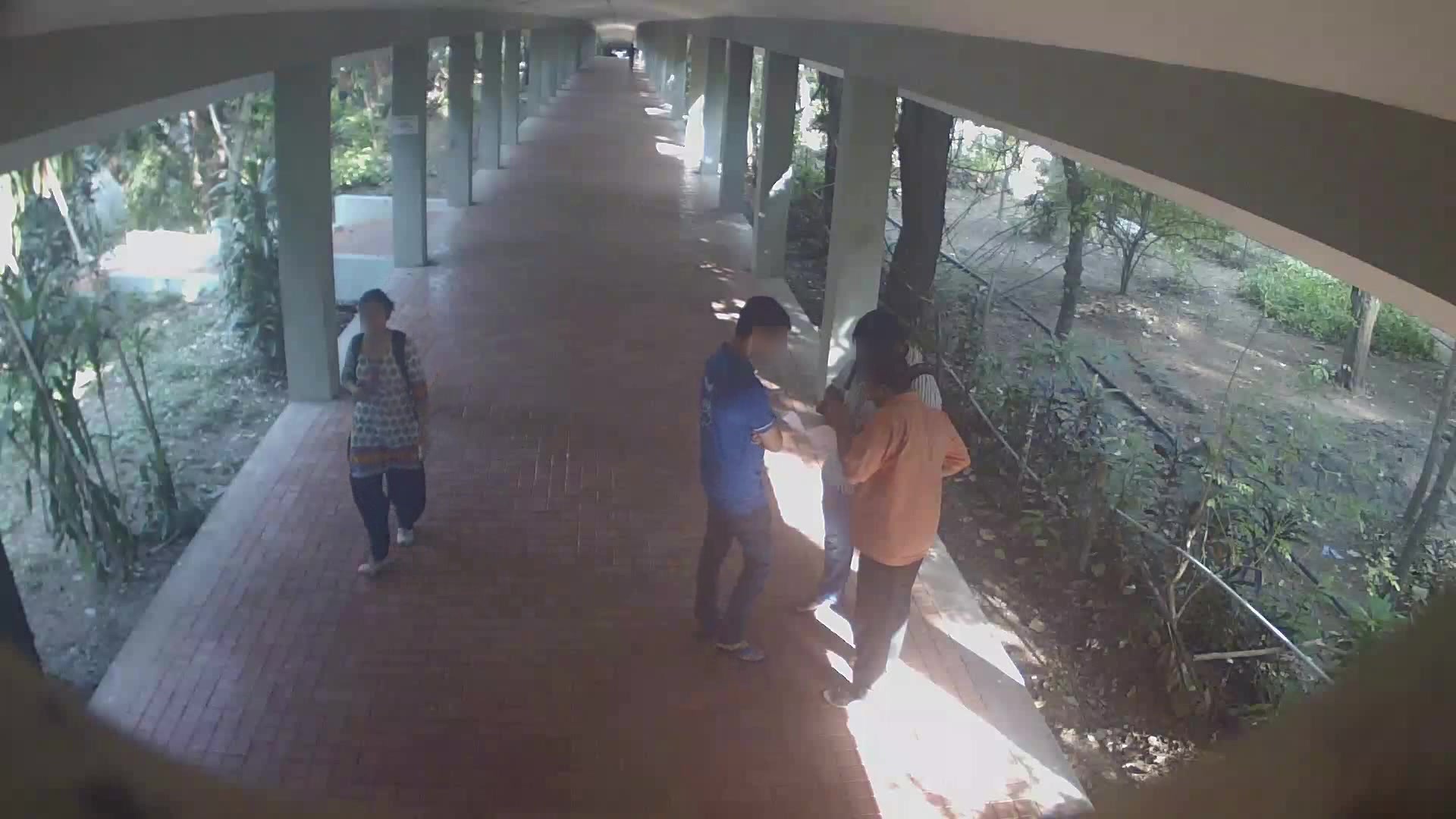}}\vspace{-0.2cm}\\
    
    \caption{Samples images from the proposed \textit{Corridor} dataset.}
    \label{fig:iitb}
\end{figure*}
\begin{table*}
\centering
\def\arraystretch{1.2}
\begin{tabular}{ | c | c | c | c | } 
 \hline
  \textbf{Dataset} &  \textbf{Training frames} & \textbf{Testing frames} & \textbf{Abnormal activities}\\ 
  \hline
 ShanghaiTech \cite{TSC-RNN} & 2,74,515 & 42,883 & Throwing Object, Jumping, Pushing\\
  & & & Riding a Bike, Loitering, Climbing\\
 \hline
 USC Ped-1 \cite{li2013anomaly} & 6,800 & 7,200
 & Bikers, small carts, walking across walkways\\
 \hline
 USC Ped-2 \cite{li2013anomaly} & 2,550 & 2,010 
 & Bikers, small carts, walking across walkways\\
 \hline
 Subway \cite{adam2008robust} & 20,000 & 116,524 & Climbing over fence, wrong direction\\
 \hline
 Avenue \cite{Avenue}&  15,328 & 15,324 & Running, Throwing object, Wrong direction\\
 \hline
 \textit{Corridor} & 3,01,999 &  1,81,567 & {Protest, Unattended Baggage, Cycling,} \\
 (proposed) & & & { Sudden Running, Fighting, Chasing, Loitering,}\\
 & & & {Suspicious Object,Hiding, Playing with Ball} \\
 \hline

\end{tabular}
\vspace{10pt}
\caption{ \textit{Corridor} (proposed) dataset compared to existing single camera datasets. This dataset is more challenging as can be seen in the Table \ref{table:performance} showing performance of state-of-the-art methods on different datasets.}
\label{table:iitb}
\end{table*}
training epoch consists of sub-epochs. In each sub-epoch, we train up to a particular layer corresponding to one of the timescales. In the first sub-epoch, we train only the first layer corresponding to timescale of 3. In the last sub-epoch, we train the complete model up to layer 7 that corresponds to timescale of 25. There are 4 sub-epochs corresponding to each timescale. To train the first sub-epoch, we split the training pose trajectories in to smaller trajectories of length 6 (3 in and 3 out). To train the last sub-epoch, we split the input trajectories in to length of 50 (25 in and 25 out). The loss after an epoch is equal to the loss incurred at the last sub-epoch. We used \textit{Adam} as the optimiser to train the model.

\subsection{Testing}
To test an input sequence, we split the input sequence in to smaller sequences of length 6, 10, 26, and 50. We use sequences of length 6 to generate predictions from layer 1 (\ie, timescale of 3). Similarly, we use sequences of lengths 10, 26, and 50 to produce predictions at layer 2, 4, and 7, respectively. This is done for both future and past prediction models. We combine the prediction errors by voting. Finally at any time instant, if the error value is higher than a threshold, it is considered as abnormal.

\subsection{Performance Evaluation}
We compare our results with \cite{TSC-RNN}, \cite{future}, \cite{hasan2016}, \cite{unmask}, \cite{novelty}, and \cite{pose}. The method proposed by Luo \etal in \cite{TSC-RNN} is based on learning a dictionary for the normal scenes under a sparse coding scheme. To smoothen the predictions over time, they proposed a Temporally-coherent Sparse Coding (TSC) formulation. Liu \etal \cite{future} proposed a future frame prediction based method to detect anomaly. They also use optical flow to enforce the temporal constraint along with the spatial closeness. The method proposed by Morais \etal \cite{pose} uses pose trajectory under the joint framework of reconstruction and prediction. To compare with these existing approaches, we also use \textit{Frame-AUC} as the evaluating criteria. The comparison is given in Table \ref{table:performance}. Our proposed model outperforms the
\begin{table*}[]
\def\arraystretch{1.5}

\centering
\begin{tabular}{|c|c|c|c|c|c|c|}
\hline
Timescales   & \multicolumn{3}{c|}{HR-ShanghaiTech} & \multicolumn{3}{c|}{HR-Avenue} \\ \hline
             & Future    & Past     & Future+Past   & Future  & Past   & Future+Past \\ \hline
3            & 71.71     & 70.62    & 72.05         & 85.33   & 83.36  & 84.99       \\ \hline
3, 5         & 72.89     & 71.69    & 73.39         & 86.96   & 84.70  & 86.82       \\ \hline
3, 5, 13     & 74.51     & 73.39    & 75.65         & \textbf{88.29}   & \textbf{86.20}  & \textbf{88.43}       \\ \hline
3, 5, 13, 25 & \textbf{74.98}     & \textbf{74.17}    & \textbf{77.04}         & 87.37   & 85.65  & 88.33       \\ \hline

\end{tabular}

\vspace{10pt}
\caption{Effect of multiple timescales and past predictions on the overall performance of the model.}
\label{table:scale}
\end{table*}
other methods on HR-ShanghaiTech and HR-Avenue datasets. Even though our model doesn't capture the non-human anomalies (\eg car), it outperforms on ShanghaiTech dataset and provides comparable performance on Avenue dataset. The proposed dataset is more challenging as can be seen in the Table \ref{table:performance}.  

\subsection{Ablation studies}
In this section, we provide ablation studies. In particular, we discuss the effects of timescales and past prediction model.
\subsubsection{Effect of multi-timescale framework}
In this sub-section, we discuss the effect of various timescales on the performance. In order to compare, we use our trained model and then test the model with incremental combinations of timescales. Table \ref{table:scale} compares the \textit{Frame AUC} when using different timescales. In most of the cases, we see an improvement in the performance as we include more timescales. However, in case of HR-Avenue, we do not see this trend in the last row. This is because the dataset does not have long-term anomaly corresponding to timescale of 25. 

 \subsubsection{Effect of past prediction model}
To see the effect of past prediction model, we compare the performances of future prediction model, past prediction model, and combined. In Table \ref{table:scale}, we can observe from the rows that adding past prediction model improved the overall performance in most of the cases.

\section{Conclusions}
\label{con}
In this work, we developed a multi-timescale framework to capture abnormal human activities occurring at different timescales. We use this framework to predict pose trajectories in both the directions (past and future) at multiple timescales. These multi-timescale predictions are used to detect abnormal activity. Our experiments shows that the multi-timescale framework outperforms the state-of-the-art models. In addition, we also release a challenging abnormal activity data set for research use. 

{\small
\bibliographystyle{ieee}
\bibliography{egbib}
}

\end{document}